\documentclass[lettersize,journal,10 pt]{IEEEtran}

\usepackage{cite}
\usepackage{amsmath,amssymb,amsfonts}
\usepackage{algorithm}
\usepackage{algpseudocode}

\usepackage{booktabs}

\usepackage{graphicx}
\usepackage{textcomp}
\usepackage{xcolor}
\usepackage{subfigure}
\usepackage{amssymb}
\usepackage{fontenc}
\usepackage{multirow}
\usepackage{url}

\usepackage{soul}
\usepackage{xcolor}

\usepackage[bookmarks=false,hidelinks]{hyperref}

\IEEEoverridecommandlockouts
\begin{document}

	\title{\LARGE \bf Trajectory-Level Mode Guidance for Controllable Diffusion-Based Multi-Robot Motion Planning
	}

	\author{ Tianyou Yu, Shengze Cai, Chao Xu*
		\thanks{
	*Corresponding Author.
	Institute of Cyber-System and Control, College of Control Science and Engineering,
	Zhejiang University, Hangzhou 310027, China.
	E-mail: \texttt{cxu@zju.edu.cn}.
}
	}
	
	\maketitle
	\thispagestyle{empty}
	\pagestyle{empty}
	

\begin{abstract}
Motion planning often admits multiple feasible solutions, making multimodal generation valuable, particularly for flexible multi-robot coordination. Diffusion models naturally learn such trajectory distributions, yet incorporating coarse and partial trajectory priors without restricting generation remains challenging. Such priors indicate a desirable region of the solution space rather than a single solution, motivating conditioned generation that preserves multimodality. In this paper, we guide trajectory generation in the clean trajectory space and progressively incorporate trajectory priors with a timestep-dependent guidance strength. At each reverse diffusion step, the reconstructed clean trajectory provides a unified space for integrating planning costs and partial trajectory priors. Planning costs are incorporated through gradient-based refinement, while the partial prior is progressively injected at the corresponding noise levels with decreasing guidance strength. This guides generation toward the prior in early stages while gradually releasing the constraint to preserve the inherent multimodality of the diffusion model. The framework naturally extends to multi-robot planning by incorporating inter-robot collision costs. Experiments on single- and multi-robot planning tasks demonstrate controllable trajectory synthesis, diverse feasible solutions, and safe multi-agent coordination.
\end{abstract}

\section{Introduction}

Motion planning often admits multiple feasible solutions that satisfy
the same task requirements, while differing in path geometry, clearance,
or coordination strategy \cite{carvalho2025motion}. This multimodality makes it desirable to generate diverse trajectory candidates while retaining the ability to
steer the generation toward preferred solutions. However, conventional optimization-based and search-based methods \cite{yu2025online} generally produce a single deterministic solution, lacking the capacity to represent diverse candidate trajectories . 

Recently, diffusion models have emerged as a powerful framework for
learning multimodal trajectory distributions. Existing diffusion-based
planners incorporate task constraints through conditional training or
test-time cost guidance. These approaches enable flexible adaptation to
different planning requirements \cite{liu2024dipper,pmlr-v162-janner22a}. However, effectively integrating trajectory-level priors into the generative process remains challenging. In practice, such priors may originate from coarse planning procedures or ambiguous human instructions. They often provide only rough, partial, or even inaccurate guidance. Nevertheless, they can identify a desirable region of the trajectory distribution. This naturally motivates a conditional generation perspective, where multimodal generation is steered toward preferred modes rather than constrained to a single solution. The key challenge is to exploit such imperfect priors while preserving the diversity of feasible trajectories. This is particularly important in multi-robot planning \cite{pmlr-v267-liang25e,shaoul2025multi,ding2025swarmdiff}, where multiple coordination patterns may coexist, yet more efficient and shorter trajectories are often desired.

In this paper, we guide trajectory generation in the clean trajectory space and progressively incorporate trajectory priors with a timestep-dependent guidance strength. At each reverse diffusion step, the reconstructed clean trajectory provides a unified space for integrating planning costs and partial trajectory priors. Planning costs are incorporated through gradient-based refinement, while the partial prior is progressively injected at the corresponding noise levels with decreasing guidance strength. This guides generation toward the prior in early stages while gradually releasing the constraint to preserve the inherent multimodality of the diffusion model. The same formulation naturally extends to multi-robot planning by incorporating inter-robot collision costs during sampling. We evaluate the proposed framework on single- and multi-robot planning tasks. Results demonstrate controllable trajectory generation, preservation of diverse feasible solutions, and effective collision-free coordination under increasingly challenging planning scenarios.

Our main contributions are summarized as follows:
\begin{itemize}
	\item We propose a clean-space guidance framework for diffusion-based
	trajectory planning, where both planning costs and trajectory priors
	are applied to the reconstructed clean trajectory rather than to noisy
	diffusion states. This provides a unified space for incorporating
	different forms of planning guidance.
	
	\item We introduce a timestep-dependent trajectory prior injection
	mechanism that maps the partial prior to the corresponding diffusion
	noise levels for progressive integration into the reverse process.
	This enables controllable prior-based steering while preserving the
	multimodality of diffusion-based trajectory generation.
	
	\item We extensively evaluate the proposed framework on single- and
	multi-robot planning tasks, demonstrating improved trajectory
	controllability and effective collision-free coordination under
	increasingly challenging planning scenarios. The dataset generation pipeline and implementation code will be publicly released upon publication.
\end{itemize}

\section{Related Work}
Trajectory planning has evolved from explicit optimization and mathematical
modeling toward learning-based approaches, offering increasing flexibility in representing diverse feasible solutions. However, incorporating task conditions and trajectory-level guidance while maintaining multimodality remains challenging, especially in multi-robot settings with coupled trajectories and inter-robot
collision constraints. In the following, we review related work from
three perspectives: constrained trajectory planning, denoising diffusion
models, and multi-robot motion planning.

\subsection{Constrained Trajectory Planning}

Given task conditions such as goals, velocity requirements, and human
interactions, traditional planning methods typically formulate a
constrained optimization problem or design heuristic functions
~\cite{yu2025online}, yielding deterministic solutions. Model predictive
control (MPC) is a representative framework, where system dynamics are
modeled as differential-equation constraints, obstacle avoidance as
inequality constraints, and human interactions are incorporated through
task-specific cost functions~\cite{schwarting2017parallel,marcano2020review}. Such explicit mathematical modeling enables precise and reliable control. However, this
precision also makes it difficult to incorporate ambiguous or
underspecified instructions, as informal guidance is often difficult to
translate into precise mathematical constraints or cost functions.

Recently, learning-based methods, such as conditioned imitation learning
~\cite{loquercio2021learning}, have shown that without modeling multiple
hypotheses, imitation learning tends to regress toward the mean of
multiple feasible behaviors, which can easily lead to failure. Such
methods avoid the need for detailed mathematical modeling but rely on
task-specific training data. Methods such as conditional variational
autoencoders (CVAEs) further introduce probabilistic latent representations to model diverse trajectory distributions, yet their
ability to capture the inherent multimodality of trajectory generation
remains limited \cite{carvalho2025motion} .

\subsection{Denoising Diffusion Models}
Denoising diffusion models~\cite{ho2020denoising} have emerged as a
powerful generative framework for learning complex data distributions,
with a particular advantage in modeling multimodal distributions through
stochastic iterative denoising. They have achieved remarkable success in high-quality image generation ~\cite{peebles2023scalable} and have been extended to 3D content generation~\cite{chu2023diffcomplete,reed2024scenesense}. Beyond visual generation, diffusion models have also been explored in planning as learned priors or auxiliary models~\cite{reed2025online}. 

With appropriate conditions or guidance signals, diffusion models can
steer generation toward different plausible solutions while preserving
distributional diversity. This property has motivated their application
to trajectory prediction~\cite{jiang2023motiondiffuser} and motion
planning~\cite{carvalho2025motion,jiang2023motiondiffuser,kumar2026neuralpathlite},
where multiple feasible solutions may naturally coexist. However,
learning a multimodal trajectory distribution alone does not provide
explicit control over preferred modes during generation. Effectively
steering such distributions using additional trajectory-level
information remains an important challenge, particularly for multi-robot
coordination, where trajectories must satisfy environmental constraints,
inter-robot collision avoidance, dynamic feasibility, and available
trajectory priors~\cite{shaoul2025multi}.

\subsection{Multi-Robot Motion Planning}
Multi-robot motion planning is challenging due to the coupled nature of
multiple trajectories, which must satisfy environmental constraints,
dynamic feasibility, and inter-robot collision avoidance. Search- and
optimization-based methods~\cite{zhou2022swarm} explicitly enforce these
constraints but typically produce deterministic solutions and require
carefully designed objectives. Multi-Agent Path Finding (MAPF) \cite{shaoul2025multi} formulates coordination as discrete constraint search, e.g., through Conflict-Based Search (CBS), but is less suited to continuous trajectories, dynamic constraints, and partial trajectory priors. Learning-based approaches~\cite{riviere2020glas,chen2023transformer} learn coordinated behaviors from data, with optimal planning methods often used to generate expert demonstrations. Reinforcement learning ~\cite{wang2020mobile} provides another alternative, but sparse rewards in large environments make effective exploration difficult, often requiring additional global guidance to direct the policy toward
promising solutions. Diffusion-based methods~\cite{shaoul2025multi} exploit multimodal generation to represent diverse coordination patterns. Notably, the
model can be trained solely on single-robot trajectories to learn a trajectory prior, which can then be transferred to multi-robot planning at inference time, making the approach convenient to deploy. Nevertheless, controlling the generated modes using coarse or partial trajectory priors remains challenging, especially when the priors may be incomplete or inaccurate.

Motivated by these limitations, we propose a diffusion-based trajectory
planning framework that incorporates planning costs and partial trajectory
priors in a unified clean trajectory space, enabling controllable mode
selection while preserving the multimodality of trajectory generation.
The framework is trained solely on single-agent trajectory data, while
multiple agent trajectories can be generated jointly in a single
inference process, enabling efficient transfer to multi-robot planning.

\section{Problem Formulation}
\label{sec:problem}

Given a diffusion model trained to capture the distribution of feasible
robot trajectories, we formulate trajectory planning as conditional
trajectory generation. Let $N$ denote the number of trajectory instances
generated in a single inference, $T$ the trajectory horizon, and $d$ the
state dimension. The generated trajectories are represented as
$\mathbf{X}\in\mathbb{R}^{N\times T\times d}$, where
$\mathbf{x}^{i}_{t}\in\mathbb{R}^{d}$ denotes the state of the
$i$-th trajectory at timestep $t$. The corresponding start and goal
states are denoted by $\mathbf{S},\mathbf{G}\in\mathbb{R}^{N\times d}$,
where $\mathbf{s}^{i}$ and $\mathbf{g}^{i}$ specify the start and goal
states of the $i$-th trajectory. 

When an additional trajectory prior
$\mathbf{X}^{\mathrm{prior}}\in\mathbb{R}^{N\times T\times d}$
is available, it provides auxiliary guidance toward a desired solution.
Since the prior may be incomplete or suboptimal, it is not treated as a
hard constraint. The objective is to generate feasible trajectories
conditioned on the task constraints and, when available, remain
consistent with the trajectory prior:
\begin{equation}
\mathbf{X}^{*}\sim
p_{\theta}\left(
\mathbf{X}\mid
\mathbf{S},\mathbf{G},
\mathcal{O},
\mathbf{X}^{\mathrm{prior}}
\right),
\label{eq:problem_formulation}
\end{equation}
where $\mathcal{O}$ denotes the task-specific planning constraints.

In multi-robot planning, the $N$ trajectory instances can naturally
correspond to $N$ robots, with each trajectory associated with its
corresponding start and goal states. The task constraints
$\mathcal{O}$ therefore include environment-dependent and
robot-dependent requirements, such as obstacle avoidance, inter-robot
collision avoidance.

\section{Method}

\subsection{System Overview}

\begin{figure}[t]
	\centering
	\includegraphics[width=0.45\textwidth]{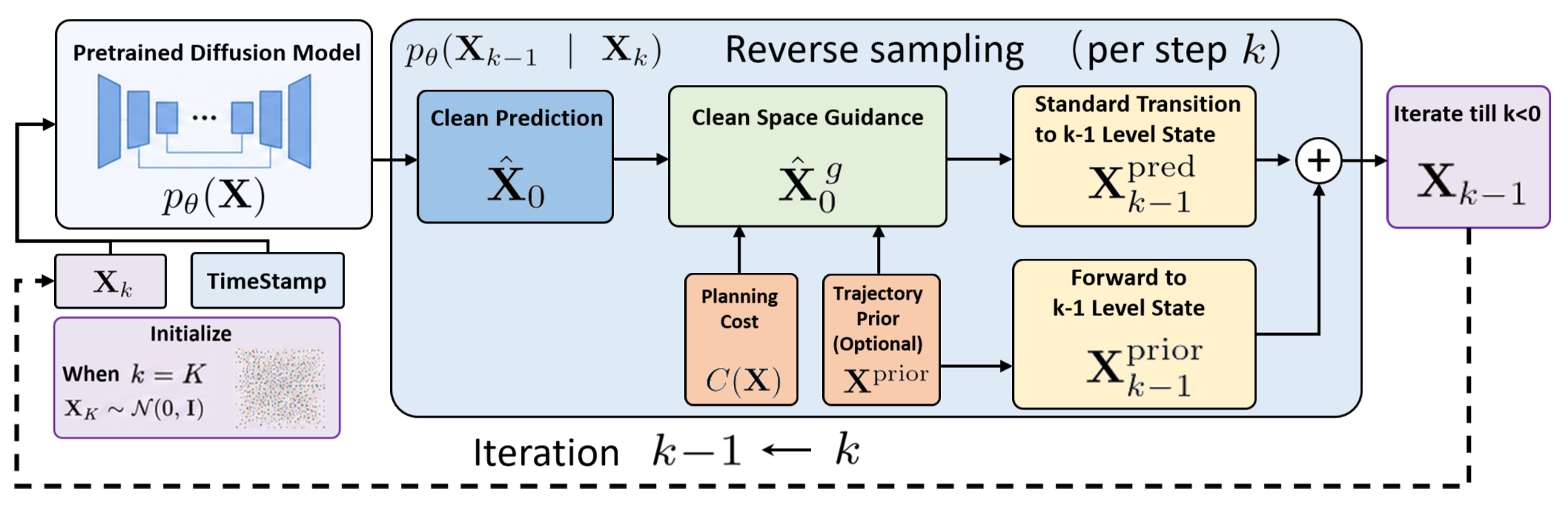}
	\caption{Overview of the proposed sampling-time guidance framework.
		The pretrained diffusion model reconstructs clean trajectories and
		incorporates planning costs and optional trajectory priors in the clean
		space during reverse diffusion, without modifying or retraining the
		model.}
	\label{fig:system}
\end{figure}

The proposed framework, illustrated in Fig.~\ref{fig:system}, uses a pretrained diffusion model as a trajectory prior, denoted by $p_{\theta}(\mathbf{X})$. Starting from $\mathbf{X}_K\sim\mathcal{N}(\mathbf{0},\mathbf{I})$, the model
iteratively reconstructs a clean trajectory estimate
$\hat{\mathbf{X}}_0$ during reverse diffusion. Task-specific
information is incorporated only at sampling time rather than as
network conditions. Specifically, the reconstructed trajectory is
guided by the planning cost and an optional trajectory prior, yielding
$\hat{\mathbf{X}}_0\rightarrow\hat{\mathbf{X}}_0^{\,g}$. The guided
clean trajectory is then mapped to the preceding diffusion state,
while the prior is additionally transformed to the corresponding noise
level for soft injection. Thus, each reverse step can be summarized as
\begin{equation}
\mathbf{X}_k
\rightarrow
\hat{\mathbf{X}}_0
\rightarrow
\hat{\mathbf{X}}_0^{\,g}
\rightarrow
\mathbf{X}_{k-1},
\end{equation}
where $\hat{\mathbf{X}}_0^{\,g}$ jointly incorporates task cost and
trajectory-prior guidance. This sampling-time formulation preserves the multimodality of $p_{\theta}(\mathbf{X})$ while enabling controllable guidance without modifying or retraining the diffusion model.

\subsection{Trajectory Diffusion Model}

We first formulate a standard diffusion model for trajectory generation,
which learns a trajectory prior from the training data. The model
consists of a forward noising process and a reverse denoising process
that map trajectories between the data and Gaussian noise distributions.

\paragraph{Forward Diffusion.}
Starting from a clean trajectory $\mathbf{X}_0$, Gaussian noise is
progressively added over $K$ diffusion steps. The forward diffusion
process is defined as a Markov chain whose joint conditional
distribution is factorized as

\begin{equation}
q(\mathbf{X}_{1:K}|\mathbf{X}_0)
=
\prod_{k=1}^{K}
q(\mathbf{X}_k|\mathbf{X}_{k-1}),
\label{eq:forward_joint}
\end{equation}
the transition at each diffusion step is defined as

\begin{equation}
q(\mathbf{X}_k|\mathbf{X}_{k-1})
=
\mathcal{N}
\left(
\mathbf{X}_k;
\sqrt{\alpha_k}\mathbf{X}_{k-1},
(1-\alpha_k)\mathbf{I}
\right),
\end{equation}
where $\alpha_k=1-\beta_k$ and $\beta_k$ denotes the noise schedule.
By defining $\bar{\alpha}_k=\prod_{j=1}^{k}\alpha_j$, the noisy trajectory at an arbitrary timestep $k$ can be directly sampled from the clean trajectory as
\begin{equation}
\mathbf{X}_k
=
\sqrt{\bar{\alpha}_k}\mathbf{X}_0
+
\sqrt{1-\bar{\alpha}_k}\boldsymbol{\epsilon},
\qquad
\boldsymbol{\epsilon}\sim\mathcal{N}(\mathbf{0},\mathbf{I}).
\label{eq:forward_sample}
\end{equation}
As $k$ increases, the trajectory gradually loses its original structure
and approaches a Gaussian distribution. A neural network
$\epsilon_{\theta}$ is trained to predict the noise
$\boldsymbol{\epsilon}$ from the noisy trajectory $\mathbf{X}_k$ and the
diffusion timestep $k$:

\begin{equation}
\mathcal{L}_{\mathrm{diff}}
=
\mathbb{E}_{\mathbf{X}_0,k,\boldsymbol{\epsilon}}
\left[
\left\|
\boldsymbol{\epsilon}
-
\epsilon_{\theta}(\mathbf{X}_k,k)
\right\|_2^2
\right].
\end{equation}
Equ.~\eqref{eq:forward_sample} provides a closed-form expression for
the noisy trajectory at an arbitrary diffusion timestep. Therefore,
the reverse transition
$p_{\theta}(\mathbf{X}_{k-1}\mid\mathbf{X}_k)$
can be constructed from the estimated clean trajectory. Specifically,
the trained network $\epsilon_{\theta}$ first predicts the noise in
$\mathbf{X}_k$, which gives an estimate of the clean trajectory:
\begin{equation}
\hat{\mathbf{X}}_0
=
\frac{
	\mathbf{X}_k
	-
	\sqrt{1-\bar{\alpha}_k}\,
	\epsilon_{\theta}(\mathbf{X}_k,k)
}{
	\sqrt{\bar{\alpha}_k}
}.
\label{eq:x0_prediction}
\end{equation}
The estimated clean trajectory $\hat{\mathbf{X}}_0$ is then used in place of
the unknown $\mathbf{X}_0$ to construct $p_{\theta}(\mathbf{X}_{k-1}\mid\mathbf{X}_k)$, which can be interpreted as a
plug-in approximation \cite{ho2020denoising}. In DDPM, the transition is stochastic and can be obtained by sampling from the corresponding Gaussian posterior $q(\mathbf{X}_{k-1}\mid\mathbf{X}_k,\mathbf{X}_0)$, with $\mathbf{X}_0$ replaced by $\hat{\mathbf{X}}_0$ (see Appendix). Thus, each reverse step implements the transition
$p_{\theta}(\mathbf{X}_{k-1}\mid\mathbf{X}_k)$, and the process is iteratively applied as $\mathbf{X}_K\rightarrow\mathbf{X}_{K-1}\rightarrow\cdots\rightarrow\mathbf{X}_0$.


\subsection{Cost-Guided Planning}

Let $\mathcal{O}$ denote the task constraints. The task-conditioned
trajectory distribution follows the Markov factorization
\begin{equation}
p(\mathbf{X}_0\mid\mathcal{O})
=
p(\mathbf{X}_N\mid\mathcal{O})
\prod_{k=1}^{N}
p(\mathbf{X}_{k-1}\mid\mathbf{X}_k,\mathcal{O}).
\label{eq:markov_factorization}
\end{equation}

At each reverse step, the task-conditioned transition satisfies
\begin{equation}
p(\mathbf{X}_{k-1}\mid\mathbf{X}_k,\mathcal{O})
\propto
p_{\theta}(\mathbf{X}_{k-1}\mid\mathbf{X}_k)
p(\mathcal{O}\mid\mathbf{X}_{k-1}),
\label{eq:conditional_transition}
\end{equation}
where $p_{\theta}(\mathbf{X}_{k-1}\mid\mathbf{X}_k)$ is parameterized
by the posterior $q(\mathbf{X}_{k-1}\mid\mathbf{X}_k,\hat{\mathbf{X}}_0)$.
This also implies that, given an arbitrary clean trajectory
$\mathbf{X}_0$, the forward diffusion process can generate a
corresponding noisy trajectory $\mathbf{X}_k$ at any prescribed noise
level according to the predefined noise schedule. 

Inspired by recent completion methods in computer vision that
incorporate clean observations into the denoising process, either by
replacing the corresponding noisy regions with appropriately noised
clean data using masking strategies~\cite{lugmayr2022repaint}, or by
guiding the denoising process through the reconstructed clean data
space~\cite{zhu2023denoising,wang2022zero}, we note that trajectory planning
primarily concerns the quality and feasibility of the clean trajectory
$\mathbf{X}_0$. In contrast, applying the planning cost directly to
highly noisy intermediate states may provide limited meaningful
guidance. 

Therefore, we perform cost guidance in the reconstructed clean
trajectory space and subsequently use the guided estimate to construct
the preceding noisy state. Specifically, rather than introducing an
additional learned guidance network, we explicitly model the task
likelihood using a planning cost $C$ \cite{huang2023diffusion}, which is defined in
Section~\ref{subsec:cost_model}:
\begin{equation}
p(\mathcal{O}\mid\hat{\mathbf{X}}_0)
\propto
\exp\!\left[-C(\hat{\mathbf{X}}_0)\right],
\label{eq:cost_likelihood}
\end{equation}
where $\hat{\mathbf{X}}_0$ serves as a plug-in estimate of the unknown
clean trajectory $\mathbf{X}_0$ (see Appendix). This explicit likelihood
formulation enables cost guidance directly through the gradient
$\nabla_{\hat{\mathbf{X}}_0}C(\hat{\mathbf{X}}_0)$, which provides a local
direction for modifying $\hat{\mathbf{X}}_0$ toward higher task
likelihood, without requiring additional training. In our framework,
this cost-based refinement is jointly performed with trajectory-prior
guidance, such that both task feasibility and prior information are
incorporated in the same reconstructed clean trajectory space.

\subsection{Trajectory Guidance}
\label{subsec:traj_guidance}

Diffusion models naturally capture multimodal trajectory distributions,
enabling diverse feasible solutions. To exploit available coarse
trajectory priors, we incorporate them directly into the reconstructed
clean trajectory, guiding generation toward desirable solutions while
preserving multimodality.

Let $\mathbf{X}^{\mathrm{prior}}$ denote a normalized prior trajectory
and $\mathbf{M}\in\{0,1\}^{2\times T}$ a binary trajectory mask, where
masked waypoints are guided by the prior, while unmasked waypoints remain
fully determined by the diffusion generation. The prior is incorporated
through a timestep-dependent interpolation in the reconstructed clean
trajectory space, rather than being provided as an additional network
condition or imposed directly on the network output.

\subsubsection{Denoised Trajectory Guidance}

At each reverse diffusion step, the model first predicts the
noise and reconstructs an estimate of the clean trajectory
$\hat{\mathbf{X}}_0$. We then gradually introduce the prior according to a
time-dependent weight $w_t$:
\begin{equation}
\hat{\mathbf{X}}_{0}^{\mathrm{prior}}
=
w_t\mathbf{X}^{\mathrm{prior}}
+
(1-w_t)\hat{\mathbf{X}}_0.
\end{equation}

The weight is defined as
\begin{equation}
w_t
=
\frac{1}{2}
\frac{
	\exp(\lambda \tau_t)-1
}{
	\exp(\lambda)-1
},
\qquad
\tau_t=\frac{t}{K},
\end{equation}
where $K$ is the total number of diffusion steps and $\lambda$ controls
the guidance schedule. The weight $w_t$ increases monotonically and
convexly from $0$ to $0.5$. We set $\lambda=2$ in all experiments.

The masked trajectory is then obtained as
\begin{equation}
\hat{\mathbf{X}}_{0,\mathrm{mask}}
=
\mathbf{M}\odot
\hat{\mathbf{X}}_{0}^{\mathrm{prior}}
+
(\mathbf{1}-\mathbf{M})\odot\hat{\mathbf{X}}_0.
\end{equation}
The resulting trajectory is subsequently refined by the planning cost
in the same clean trajectory space:
\begin{equation}
\hat{\mathbf{X}}_0^{\mathrm{g}}
=
\hat{\mathbf{X}}_{0,\mathrm{mask}}
-
\eta_k
\nabla_{\hat{\mathbf{X}}_{0,\mathrm{mask}}}
C\!\left(
\hat{\mathbf{X}}_{0,\mathrm{mask}}
\right).
\label{eq:cost_guidance}
\end{equation}
Thus, the trajectory prior and planning cost are jointly applied in the
reconstructed clean trajectory space before the guided trajectory is
mapped back to the diffusion space.

\subsubsection{Noisy-State Trajectory Guidance}
\label{subsubsec:noisy_traj_guidance}
The clean-space guided trajectory $\hat{\mathbf{X}}_0^{\mathrm{g}}$
is then mapped back to the diffusion space through the standard reverse
transition, $q(\mathbf{X}_{k-1}\mid\mathbf{X}_k,\hat{\mathbf{X}}_0^{\mathrm{g}})$,
yielding the predicted state $\mathbf{X}_{k-1}^{\mathrm{pred}}$. To further maintain the trajectory prior in the diffusion space, we construct a noisy representation of the prior at the same noise level:
\begin{equation}
\mathbf{X}_{k-1}^{\mathrm{prior}}
=
\sqrt{\bar{\alpha}_{k-1}}\mathbf{X}^{\mathrm{prior}}
+
\sqrt{1-\bar{\alpha}_{k-1}}\,
\boldsymbol{\epsilon},~\boldsymbol{\epsilon}
\sim\mathcal{N}(\mathbf{0},\mathbf{I}).
\end{equation}

The noisy prior is then softly incorporated into the candidate state
using the same timestep-dependent guidance weight $w_t$ defined above:
\begin{equation}
\mathbf{X}_{k-1}^{\mathrm{g}}
=
w_t\mathbf{X}_{k-1}^{\mathrm{prior}}
+
(1-w_t)\mathbf{X}_{k-1}^{\mathrm{pred}}.
\end{equation}
As denoising proceeds from $k=K$ to $k=1$, the prior is maintained at
the corresponding diffusion noise level, while the guidance weight
$w_t$ decreases convexly from $0.5$ to $0$. This convex schedule causes
the guidance strength to decrease rapidly at the early denoising stages,
gradually releasing the prior constraint and allowing the diffusion model
to recover its multimodal generation capability at an earlier stage.

Finally, the trajectory mask is applied in the noisy space:
\begin{equation}
\mathbf{X}_{k-1}
=
\mathbf{M}\odot\mathbf{X}_{k-1}^{\mathrm{g}}
+
(\mathbf{1}-\mathbf{M})
\odot\mathbf{X}_{k-1}^{\mathrm{pred}}.
\end{equation}
This process allows the partial prior to influence the denoising process
at both the clean and noisy trajectory levels, while leaving the
unmasked waypoints fully determined by the diffusion model and cost
guidance.

\subsection{Cost Function}
\label{subsec:cost_model}

The planning cost $C(\mathbf{X})$ jointly evaluates trajectory
feasibility and motion quality:
\begin{equation}
C(\mathbf{X})
=
\lambda_{\mathrm{vel}}C_{\mathrm{vel}}
+
\lambda_{\mathrm{sdf}}C_{\mathrm{sdf}}
+
\lambda_{\mathrm{goal}}C_{\mathrm{goal}}
+
\lambda_{\mathrm{col}}C_{\mathrm{col}}
+
\lambda_{\mathrm{sm}}C_{\mathrm{sm}},
\label{eq:planning_cost}
\end{equation}
where the five terms penalize excessive velocity, insufficient
obstacle clearance, deviation from the goal states, inter-robot
collisions, and trajectory irregularity, respectively.

\paragraph{Velocity Cost.}
The velocity cost penalizes violations of the maximum velocity
$v_{\max}$. The discrete velocity of robot $i$ at timestep $t$ is
$\mathbf{v}^{i}_{t}=(\mathbf{x}^{i}_{t+1}-\mathbf{x}^{i}_{t})/\Delta t$,
and the corresponding cost is
\begin{equation}
C_{\mathrm{vel}}
=
\sum_{i=1}^{N}\sum_{t=1}^{T-1}
\left[
\max\left(0,\|\mathbf{v}^{i}_{t}\|_2-v_{\max}\right)
\right]^2.
\label{eq:velocity_cost}
\end{equation}

\paragraph{SDF Cost.}
Let $D(\mathbf{x})$ denote the signed distance field of the environment
and $d_{\mathrm{safe}}$ the desired obstacle clearance. We penalize
states whose distance to the nearest obstacle falls below this margin:
\begin{equation}
C_{\mathrm{sdf}}
=
\sum_{i=1}^{N}\sum_{t=1}^{T}
\left[
\max\left(0,d_{\mathrm{safe}}-D(\mathbf{x}^{i}_{t})\right)
\right]^2.
\label{eq:sdf_cost}
\end{equation}

\paragraph{Goal Cost.}
To encourage the generated trajectories to reach their corresponding
goal states, we penalize the terminal position error:
\begin{equation}
C_{\mathrm{goal}}
=
\sum_{i=1}^{N}
\left\|
\mathbf{x}^{i}_{T}-\mathbf{g}^{i}
\right\|_2^2.
\label{eq:goal_cost}
\end{equation}

\paragraph{Multi-Robot Collision Cost.}
To enforce sufficient separation between robots, we penalize pairwise
distances below the required clearance $d_{\mathrm{robot}}$:
\begin{equation}
C_{\mathrm{col}}
=
\sum_{t=1}^{T}\sum_{i=1}^{N}\sum_{j=i+1}^{N}
\left[
\max\left(
0,d_{\mathrm{robot}}
-\|\mathbf{x}^{i}_{t}-\mathbf{x}^{j}_{t}\|_2
\right)
\right]^2.
\label{eq:collision_cost}
\end{equation}

\paragraph{Smoothness Cost.}
Trajectory smoothness is encouraged by penalizing the second-order
difference of consecutive states:
\begin{equation}
C_{\mathrm{sm}}
=
\sum_{i=1}^{N}\sum_{t=2}^{T-1}
\left\|
\mathbf{x}^{i}_{t+1}
-2\mathbf{x}^{i}_{t}
+\mathbf{x}^{i}_{t-1}
\right\|_2^2.
\label{eq:smoothness_cost}
\end{equation}


\section{Experiments}

\subsection{Implementation Details}
\label{subsec:implementation}

We use a 1D U-Net with three encoder--decoder stages and a base channel
dimension of 64 to model trajectories with $T=60$ waypoints and
$d=2$ state dimensions. Diffusion timesteps are embedded into a
256-dimensional feature space. No task-specific conditions are provided
to the network, keeping the learned trajectory distribution
task-agnostic. The training dataset contains 20,000 feasible single-robot
trajectories generated in randomly constructed environments. Collision-free
start and goal states are randomly sampled, followed by A* path planning
and trajectory optimization for smoothness and obstacle clearance. Only
trajectories with at least 60 waypoints are retained. The environment
resolution is $0.05\,\mathrm{m}$ with a temporal interval of
$\Delta t=0.1\,\mathrm{s}$, which is consistently used throughout
training and inference. Since trajectory samples have different starting positions, each trajectory is translated to a local coordinate frame and normalized by a fixed spatial scale of $10\,\mathrm{m}$. During inference, the reconstructed clean
trajectory is transformed back to the original map coordinate system before
cost evaluation.

The diffusion model is trained for 150K iterations using the standard
noise-prediction objective with Adam, a batch size of 128, and a learning
rate of $1\times10^{-4}$. The diffusion process uses $K=1000$ timesteps.
DDIM is employed for accelerated sampling, with 300 sampling steps by
default unless otherwise specified. During inference, cost guidance is
applied to the reconstructed clean trajectory using the planning cost
defined in Eq.~\ref{eq:planning_cost}. The weights
$(\lambda_{\mathrm{vel}},\lambda_{\mathrm{sdf}},\lambda_{\mathrm{goal}},
\lambda_{\mathrm{col}},\lambda_{\mathrm{sm}})$ are set to
$(0.2,0.1,1.0,0.25,0.002)$, respectively. We use
$v_{\max}=1.0\,\mathrm{m/s}$ and
$d_{\mathrm{safe}}=d_{\mathrm{robot}}=1.0\,\mathrm{m}$, with a
cost-guidance step size of $\eta=0.0025$.

\subsection{Effect of Partial Trajectory Conditioning}

To investigate the effect of partial trajectory conditioning on the
generated trajectory distribution, we first evaluate the proposed
framework in a single-agent setting. Specifically, we retain $N$
trajectory instances generated in a single inference, while assigning
the same start and goal states to all instances, i.e.,
$\mathbf{s}^{1}=\cdots=\mathbf{s}^{N}$ and
$\mathbf{g}^{1}=\cdots=\mathbf{g}^{N}$. Since these trajectory instances
represent multiple candidate solutions for the same single-agent
planning problem rather than different robots, the inter-robot collision
cost is set to $C_{\mathrm{col}}=0$.  This setting provides a clearer view
of how partial trajectory priors influence the diffusion generation
process while preserving its inherent multimodal nature.

We consider two strategies for providing partial trajectory conditions: \textit{Prefix-Ratio} and \textit{Uniform-Sparse}. Given a start and a goal position, we first construct a trajectory prior by directly connecting the two positions and uniformly dividing the line segment into a fixed number of waypoints. For \textit{Prefix-Ratio}, the first $r\%$ of the prior waypoints are provided as the trajectory mask, where $r$ denotes the conditioning ratio. For \textit{Uniform-Sparse}, a fixed number of waypoints are uniformly selected along the entire prior trajectory and provided as the mask. Unlike \textit{Prefix-Ratio}, these conditions are distributed across the trajectory and therefore provide sparse information about the overall trajectory shape.

\begin{figure}[t]
	\centering
	\includegraphics[width=0.5\textwidth]{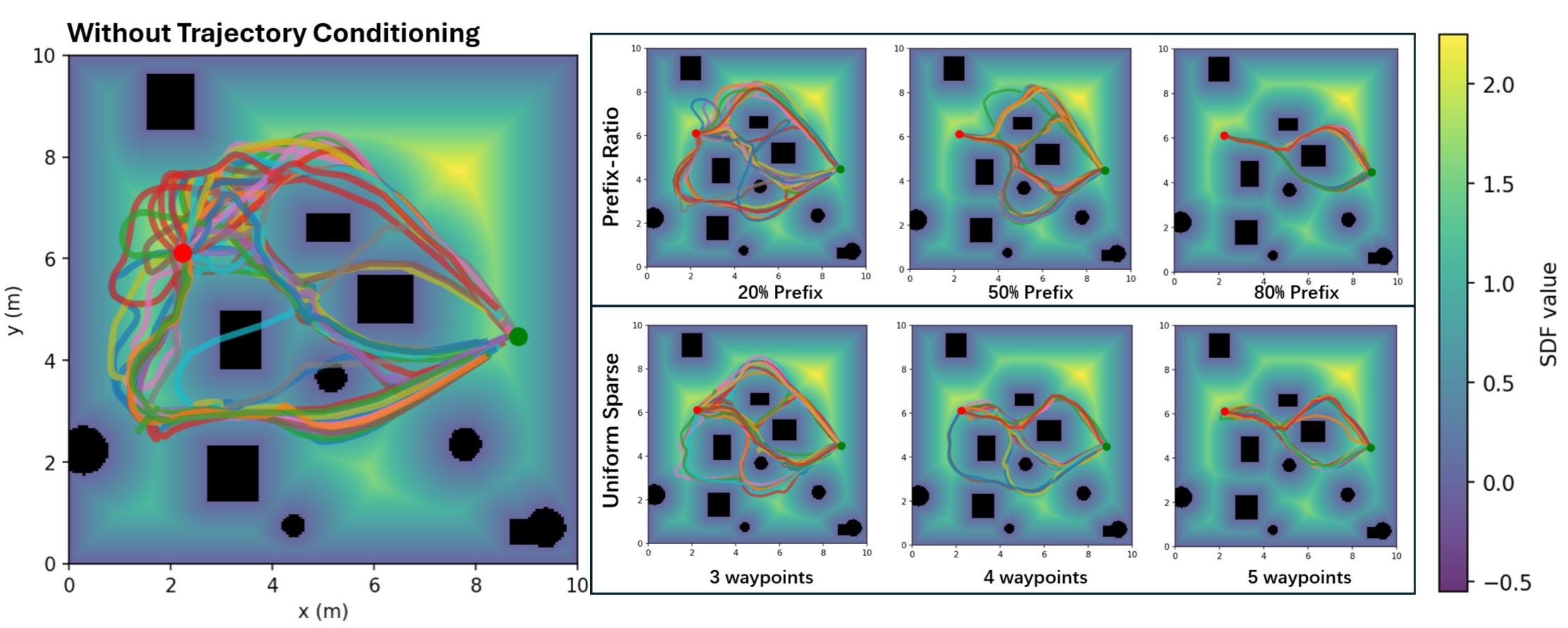}
	\caption{Single-agent trajectory generation under different partial trajectory conditioning strategies in an SDF environment. We consider \textit{Prefix-Ratio} and \textit{Uniform-Sparse} conditions. For each condition, 64 trajectories are generated in a single inference batch and visualized using different colors. The red and green circles denote the start and goal positions, respectively.}
	\label{fig:mask_traj}
\end{figure}

As shown in Fig.~\ref{fig:mask_traj}, the amount of trajectory conditioning has a clear effect on the diversity of the generated trajectories. Without any conditioning, the diffusion model produces a relatively diverse set of feasible trajectories, indicating that the learned trajectory distribution contains multiple possible solutions between the given start and goal. After introducing partial trajectory conditions, the generated trajectories become progressively more concentrated around the provided prior. In particular, increasing the prefix ratio in \textit{Prefix-Ratio}, or increasing the number of conditioned waypoints in \textit{Uniform-Sparse}, progressively reduces the diversity of the generated samples.

This observation demonstrates that partial trajectory conditioning can guide the diffusion model while preserving its inherent multimodal generation capability. Without conditioning, the model explores a broad range of feasible trajectory modes. As more prior waypoints are provided, the generated trajectories become progressively concentrated around the prior trajectory, while still maintaining multiple distinct solutions. Thus, the proposed masking mechanism does not collapse the diffusion model to a deterministic solution, but instead provides a controllable way to trade off solution diversity and prior guidance.

\subsection{Multi-Robot Planning}
We conduct multi-robot trajectory planning experiments with different numbers of robots using our proposed diffusion planning framework, in order to evaluate its effectiveness in multi-robot coordination and collision avoidance.

\subsubsection{Multi-Robot Position Swapping}

We first consider a position-swapping task in an obstacle-free environment.
The robots are initially distributed uniformly on a circle and each robot is
assigned the initial position of another robot as its goal. We evaluate
scenarios with 4, 6, 8, and 10 robots. The corresponding circle radii are
$3\,\mathrm{m}$ for the 4-, 6-, and 8-robot cases, and $4\,\mathrm{m}$ for the
10-robot case. This setup requires the robots to coordinate their motions
while avoiding inter-robot collisions. 

\begin{figure*}[t]
	\centering
	\includegraphics[width=0.79\textwidth]{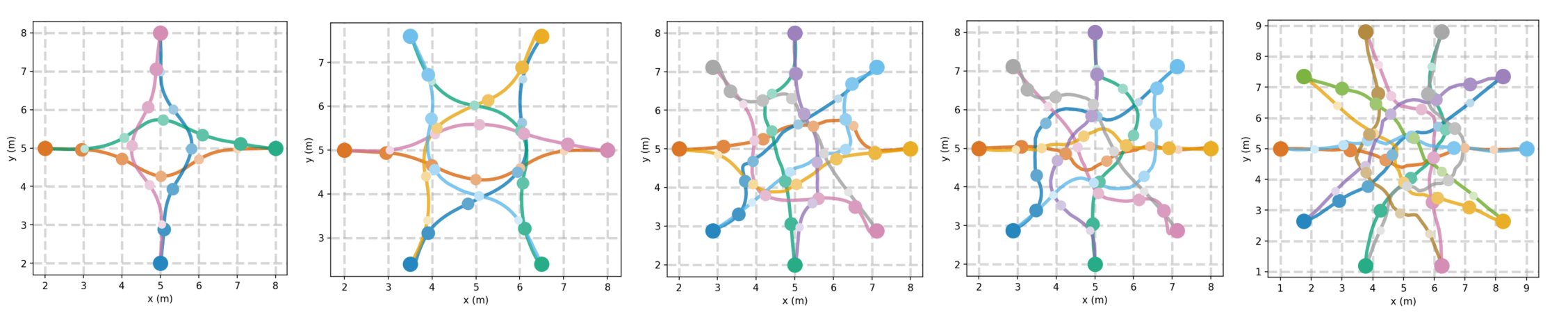}
	\caption{
		Multi-robot position-swapping results with different numbers of robots.
		From left to right, the scenarios contain 4, 6, 8, 8, and 10 robots.
		The robots are initially distributed on circles with radii of
		$3\,\mathrm{m}$ for the 4-, 6-, and 8-robot cases and
		$4\,\mathrm{m}$ for the 10-robot case, and each robot is assigned
		another robot's initial position as its goal.
		Different colors denote different robots. Circles with gradually
		increasing size and darker color indicate the temporal progression
		of each trajectory. Two different inference results are shown for the
		8-robot case to illustrate the multimodal nature of the diffusion-based
		planner.
	}
	\label{fig:ma}
\end{figure*}

Since the position-swapping task admits a relatively large feasible solution space and can therefore produce multiple valid trajectory configurations, we adopt the \textit{Uniform-Sparse} conditioning strategy, using 30 waypoints as the trajectory prior. This provides a relatively strong mode constraint while retaining sufficient freedom for the diffusion model to generate alternative feasible solutions. As shown in Fig.~\ref{fig:ma}, the proposed diffusion planning framework successfully generates collision-free trajectories with an increasing number of robots. For the 8-robot case, two different inference results are additionally presented. Although the initial and goal configurations remain unchanged, the generated trajectories exhibit different coordination patterns, demonstrating that the diffusion-based planner can preserve multiple feasible trajectory modes rather than deterministically producing a single coordination solution.

The quantitative results are summarized in Table~\ref{tab:ma}.
For all robot configurations, increasing the number of masked waypoints
generally leads to shorter generated trajectories. In particular, the
30-waypoint condition consistently achieves the lowest mean trajectory
length for 4, 6, and 8 robots, while also providing competitive results
for the 10-robot case. This indicates that a denser prior provides stronger
structural guidance, allowing the planner to generate more direct and
coordinated trajectories.

Meanwhile, the minimum inter-robot distance remains above
$0.7\,\mathrm{m}$ for all evaluated configurations, demonstrating that
the generated trajectories maintain a safe separation between robots.
Overall, these results show that providing more prior waypoints can
effectively improve trajectory efficiency while maintaining collision
avoidance, demonstrating the effectiveness of partial trajectory
conditioning in multi-robot planning.

\begin{table}[t]
	\centering
	\caption{Multi-agent trajectory planning performance for the position-swapping task in an obstacle-free environment with different numbers of masked waypoints.}
	\setlength{\tabcolsep}{4.5pt}
	\renewcommand{\arraystretch}{0.9}
	\begin{tabular}{cc|ccc|c}
		\toprule
		\multirow{2}{*}{Agents}
		& \multirow{2}{*}{Waypoints}
		& \multicolumn{3}{c|}{Trajectory Length (m)}
		& \multirow{2}{*}{Min Dist. (m)} \\
		\cmidrule(lr){3-5}
		& & Max & Mean & Min & \\
		\midrule
		
		\multirow{3}{*}{4}
		& 30 & \textbf{6.676} & \textbf{6.391} & 6.253 & 1.018 \\
		& 15 & 7.022 & 6.594 & 6.266 & 1.032 \\
		& 6  & 7.178 & 6.646 & \textbf{6.242} & \textbf{1.073} \\
		
		\midrule
		
		\multirow{3}{*}{6}
		& 30 & \textbf{6.839} & \textbf{6.558} & \textbf{6.264} & 0.955 \\
		& 15 & 7.160 & 6.726 & 6.314 & \textbf{0.972} \\
		& 6  & 8.084 & 7.277 & 6.633 & 0.964 \\
		
		\midrule
		
		\multirow{3}{*}{8}
		& 30 & \textbf{7.431} & \textbf{6.784} & \textbf{6.242} & 0.901 \\
		& 15 & 7.777 & 6.900 & 6.213 & 0.940 \\
		& 6  & 8.526 & 7.359 & 6.486 & \textbf{0.966} \\
		
		\midrule
		
		\multirow{3}{*}{10}
		& 30 & 9.504 & \textbf{8.796} & 8.188 & 0.705 \\
		& 15 & \textbf{9.432} & 8.808 & 8.265 & 0.806 \\
		& 6  & 10.014 & 8.917 & \textbf{8.231} & \textbf{0.904} \\
		
		\bottomrule
	\end{tabular}
	\label{tab:ma}
\end{table}

\subsubsection{Multi-Robot Planning in Obstacle Environments}

\begin{figure}[t]
	\centering
	\subfigure[]{\includegraphics[width=1.1in]{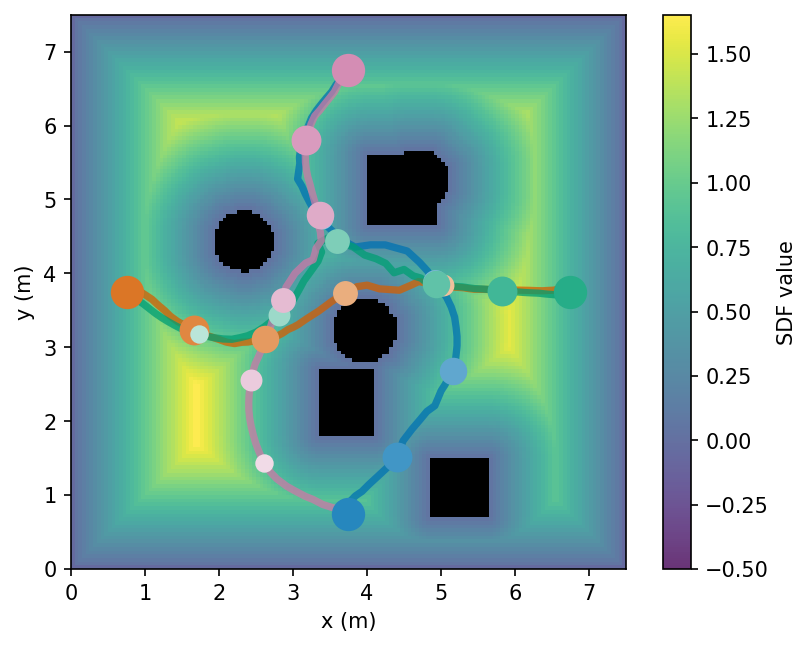}}
	\subfigure[]{\includegraphics[width=1.1in]{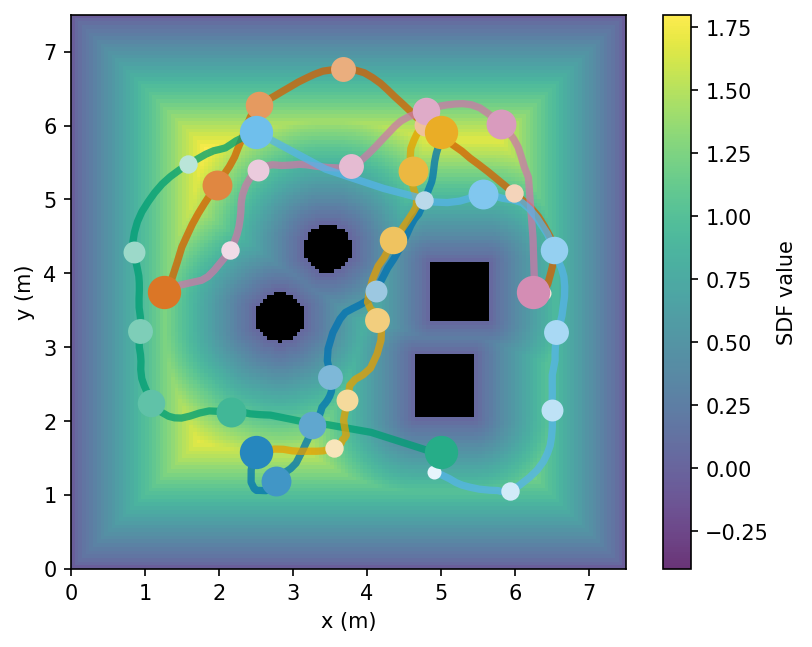}}
	\subfigure[]{\includegraphics[width=1.1in]{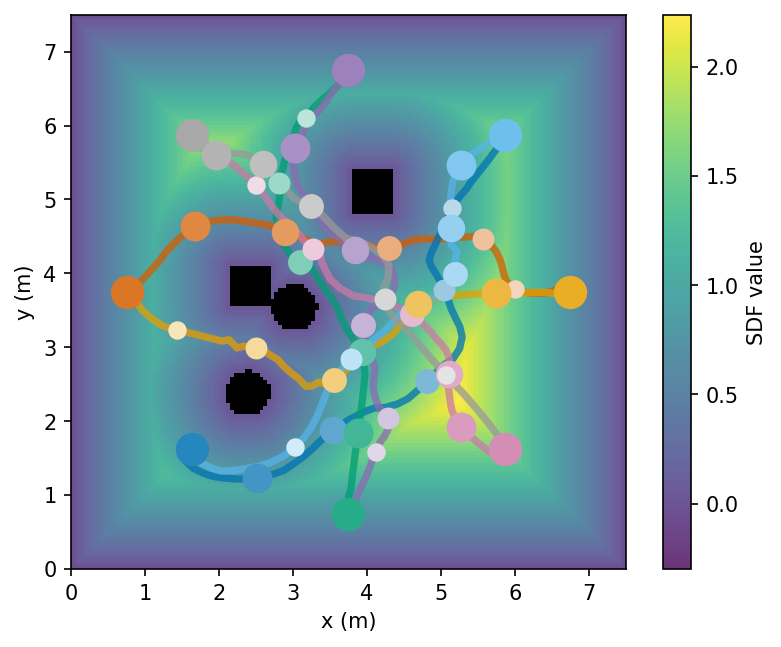}}
	\caption{   Multi-robot position-swapping in environments with static obstacles.
				The generated trajectories simultaneously satisfy the goal constraints,
				inter-robot collision avoidance, and obstacle avoidance.}
	\label{fig:ma_sdf}
\end{figure}

We further consider the more challenging case where static obstacles are introduced into the environment. In this setting, the planner must simultaneously satisfy the goal constraints, avoid collisions between robots, and avoid collisions with environmental obstacles. Since the obstacles reduce the available space for multi-robot position swapping and make the coordination problem more constrained, we adopt the \textit{Uniform-Sparse} conditioning strategy with 6 waypoints as the trajectory prior. As shown in Fig.~\ref{fig:ma_sdf}, the generated trajectories generally follow the provided trajectory prior while adapting to the environmental constraints. In particular, the presence of obstacles may make the direct position-swapping patterns observed in the obstacle-free case infeasible. In Fig.~\ref{fig:ma_sdf}(b), one pair of robots successfully performs the position swap through the central region, while the remaining robots avoid the constrained central area and exchange their positions through rotational motions along the outer region. This result demonstrates that the proposed method does not rigidly enforce the trajectory prior, but instead balances prior guidance with task-specific planning costs to adapt the generated trajectories to the available free space.

\subsubsection{Denoising Process with Prior Guidance}

To better visualize the denoising process, we increase the number of
DDIM sampling steps to 500 and visualize the reconstructed clean
trajectories $\hat{\mathbf{X}}^0$ at selected denoising steps. 

\begin{figure*}[t]
	\centering
	\includegraphics[width=0.8\textwidth]{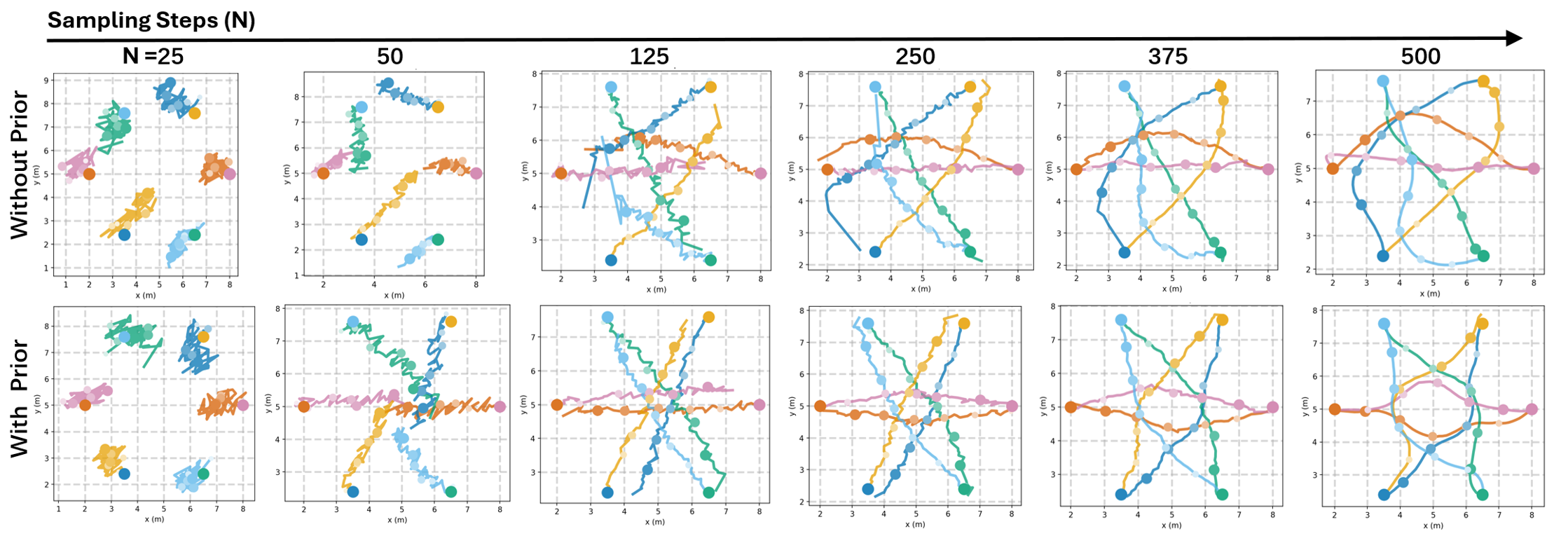}
	\caption{
		Denoising process for 6 robots performing position swapping in an
		obstacle-free environment, with and without prior guidance. The top row shows the denoising process without prior guidance, while the bottom row shows the process with the \textit{Uniform-Sparse} prior consisting of 30 waypoints. From left to right, the trajectories are shown from the beginning to the end of the denoising process. For clarity, the goal position of each agent is indicated in advance by a corresponding colored circle.}
	\label{fig:ddim_ma}
\end{figure*}

As shown in Fig.~\ref{fig:ddim_ma}, in the obstacle-free environment, without prior guidance, the trajectories are primarily guided toward their respective target positions by the goal cost while maintaining inter-robot collision avoidance. During denoising, the stochastic nature of diffusion sampling leads the trajectories to progressively develop in different spatial directions. Some agents pass through the central region, while others are driven toward the outer region to avoid potential collisions, resulting in distinct motion patterns and larger detours. In contrast, with the \textit{Uniform-Sparse} prior consisting of 30 waypoints, the denoising process receives additional structural guidance, allowing the trajectories to develop in a more consistent manner and form a regular position-swapping pattern.

\begin{figure*}[t]
	\centering
	\includegraphics[width=0.8\textwidth]{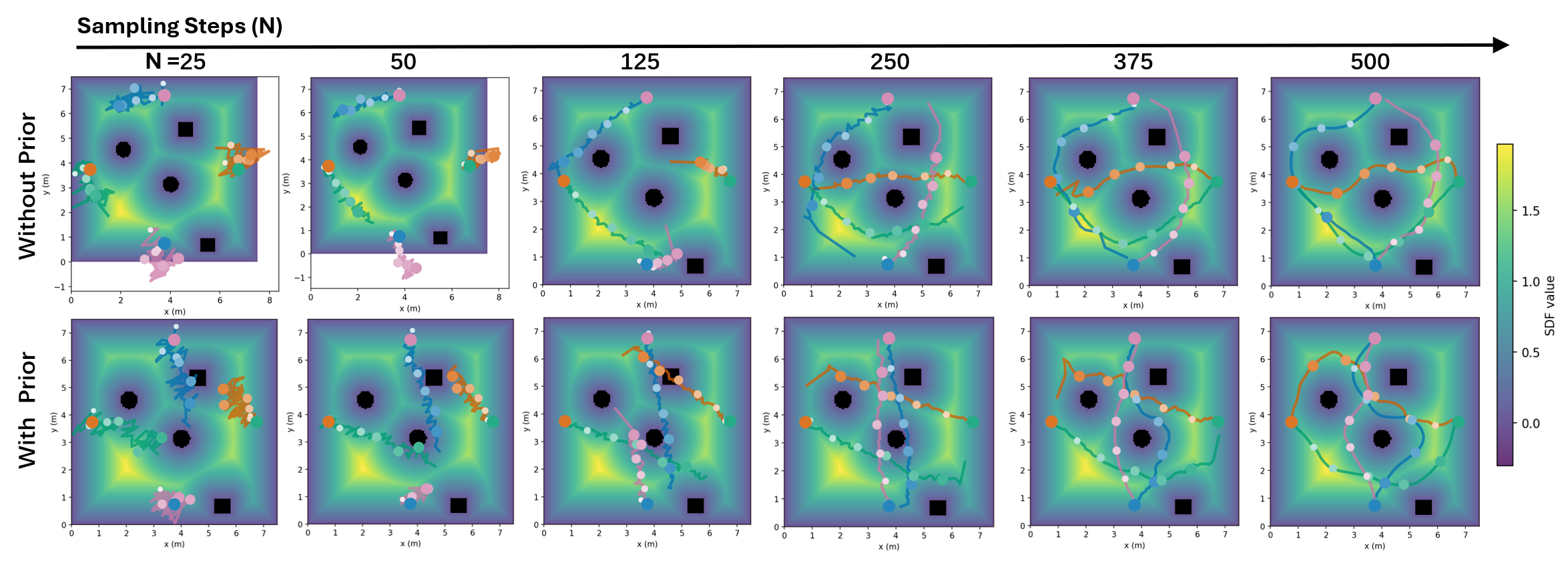}
	\caption{
		Denoising process for 4 robots performing position swapping in an
		environment with a central obstacle. The top row shows the denoising
		process without prior guidance, while the bottom row uses the \textit{Uniform-Sparse} prior consisting of 6 waypoints. 
	}
	\label{fig:ddim_ma_sdf}
\end{figure*}

In Fig.~\ref{fig:ddim_ma_sdf}, we further examine the denoising process
in an obstacle environment using a \textit{Uniform-Sparse} prior with only
6 waypoints. The partial prior provides coarse structural guidance, while
the generated trajectories are progressively refined to satisfy both
obstacle and inter-robot collision constraints. Although all six
trajectories are generated jointly within a single inference process,
distinct interaction patterns emerge during denoising. In particular, the
blue trajectory is guided toward its goal at an earlier stage, resulting
in a longer trajectory, while the green and orange trajectories gradually
detour around the obstacle. Meanwhile, the pink trajectory avoids both
the green and blue trajectories and remains relatively undeveloped during
the earlier denoising steps. It only begins to extend toward its goal
around the 125th denoising step, exhibiting an implicit waiting behavior
induced by the evolving inter-robot interactions. This behavior emerges
without explicitly modeling a waiting action or providing waiting labels,
showing that the joint denoising process can implicitly coordinate the
spatial and temporal evolution of multiple trajectories. The resulting
trajectories preserve the coarse structure of the partial prior while
adapting to environmental and inter-robot constraints.

\section{Conclusion}

We presented a diffusion-based trajectory planning framework that guides
multimodal generation in the clean trajectory space. Planning costs and
partial trajectory priors are incorporated through gradient-based guidance
and timestep-dependent prior injection, steering generation toward
preferred solutions while preserving multimodality. The framework extends
to multi-robot planning through joint trajectory generation and
inter-robot collision guidance. Experiments demonstrate controllable
generation, diverse feasible solutions, and effective collision-free
coordination. Future work will explore less structured guidance, including
ambiguous language instructions and high-level behavior labels, for more
expressive trajectory control.

\section*{Appendix}
\label{sec:appendix}
The estimated clean trajectory $\hat{\mathbf{X}}_0$ is used as a point
estimate of the unknown clean trajectory $\mathbf{X}_0$. Thus, we adopt
the following plug-in approximation:
\begin{equation}
\begin{aligned}
p_{\theta}(\mathbf{X}_{k-1}\mid\mathbf{X}_k)
&=
\int
q(\mathbf{X}_{k-1}\mid\mathbf{X}_k,\mathbf{X}_0)
p(\mathbf{X}_0\mid\mathbf{X}_k)
\,d\mathbf{X}_0
\\
&\approx_{\text{plug-in}}
\int
q(\mathbf{X}_{k-1}\mid\mathbf{X}_k,\mathbf{X}_0)
\delta\!\left(
\mathbf{X}_0-\hat{\mathbf{X}}_0
\right)
\,d\mathbf{X}_0
\\
&=
q(\mathbf{X}_{k-1}\mid\mathbf{X}_k,\hat{\mathbf{X}}_0).
\end{aligned}
\label{eq:plugin_reverse}
\end{equation}

Using the Gaussian forward process, the conditional posterior
$q(\mathbf{X}_{k-1}\mid\mathbf{X}_k,\mathbf{X}_0)$ admits a closed-form
Gaussian expression:
\begin{equation}
q(\mathbf{X}_{k-1}\mid\mathbf{X}_k,\mathbf{X}_0)
=
\mathcal{N}
\left(
\tilde{\boldsymbol{\mu}}_k,
\tilde{\beta}_k\mathbf{I}
\right),
\end{equation}
where
\begin{equation}
\tilde{\boldsymbol{\mu}}_k
=
\frac{
	\sqrt{\bar{\alpha}_{k-1}}\beta_k
}{
	1-\bar{\alpha}_k
}
\mathbf{X}_0
+
\frac{
	\sqrt{\alpha_k}(1-\bar{\alpha}_{k-1})
}{
	1-\bar{\alpha}_k
}
\mathbf{X}_k,
\end{equation}
and
\begin{equation}
\tilde{\beta}_k
=
\frac{
	1-\bar{\alpha}_{k-1}
}{
	1-\bar{\alpha}_k
}
\beta_k.
\end{equation}

Therefore, applying the plug-in estimate
$\mathbf{X}_0\approx\hat{\mathbf{X}}_0$ gives
\begin{equation}
p_{\theta}(\mathbf{X}_{k-1}\mid\mathbf{X}_k)
\approx
\mathcal{N}
\left(
\tilde{\boldsymbol{\mu}}_k(\mathbf{X}_k,\hat{\mathbf{X}}_0),
\tilde{\beta}_k\mathbf{I}
\right).
\label{eq:plugin_ddpm}
\end{equation}

The same plug-in approximation gives
\begin{equation}
\begin{aligned}
p(\mathcal{O}\mid\mathbf{X}_k)
&=
\int
p(\mathcal{O}\mid\mathbf{X}_0)
p(\mathbf{X}_0\mid\mathbf{X}_k)
\,d\mathbf{X}_0
\\
&\approx_{\text{plug-in}}
p(\mathcal{O}\mid\hat{\mathbf{X}}_0).
\end{aligned}
\label{eq:plugin_task_likelihood}
\end{equation}
With the cost-based likelihood model (Equ. \eqref{eq:cost_likelihood}), the planning cost can consequently be evaluated on the reconstructed clean trajectory.

\bibliographystyle{IEEEtran}
\bibliography{mylib}
\end{document}